\documentclass[lettersize,journal]{IEEEtran}
\usepackage{cite}
\usepackage{amsmath,amssymb,amsfonts}
\usepackage{graphicx}
\usepackage{algorithm}
\usepackage{algpseudocode}
\usepackage{multirow} 
\usepackage{textcomp}

\begin{document}
\title{CT-SDM: A Sampling Diffusion Model for Sparse-View CT Reconstruction across All Sampling Rates}
\author{Liutao Yang, Jiahao Huang, Guang Yang \IEEEmembership{Senior Member, IEEE}, Daoqiang Zhang \IEEEmembership{Senior Member, IEEE}
\thanks{L. Yang, and D. Zhang are with the College of Computer Science and Technology, Nanjing University of Aeronautics and Astronautics, China. J. Huang and G. Yang are with the Bioengineering Department and Imperial-X, Imperial College London, UK. L, National Heart and Lung Institute, Imperial College London, UK, and with National Heart Cardiovascular Research Centre, Royal Brompton Hospital, London, UK. Co-last author: Daoqiang Zhang, Guang Yang. Corresponding author: Daoqiang~Zhang. (email: dqzhang@nuaa.edu.cn.) }}
\maketitle
\begin{abstract}
Sparse views X-ray computed tomography has emerged as a contemporary technique to mitigate radiation dose. Because of the reduced number of projection views, traditional reconstruction methods can lead to severe artifacts. Recently, research studies utilizing deep learning methods has made promising progress in removing artifacts for Sparse-View Computed Tomography (SVCT). However, given the limitations on the generalization capability of deep learning models, current methods usually train models on fixed sampling rates, affecting the usability and flexibility of model deployment in real clinical settings. To address this issue, our study proposes a adaptive reconstruction method to achieve high-performance SVCT reconstruction at any sampling rate. Specifically, we design a novel imaging degradation operator in the proposed sampling diffusion model for SVCT (CT-SDM) to simulate the projection process in the sinogram domain. Thus, the CT-SDM can gradually add projection views to highly undersampled measurements to generalize the full-view sinograms. By choosing an appropriate starting point in diffusion inference, the proposed model can recover the full-view sinograms from any sampling rate with only one trained model. Experiments on several datasets have verified the effectiveness and robustness of our approach, demonstrating its superiority in reconstructing high-quality images from sparse-view CT scans across various sampling rates.
\end{abstract}


\section{Introduction}
\IEEEPARstart{I}n the fields of medical imaging, while X-ray computed tomography (CT) is widely used for its ability to generate high-quality and detailed images, it also raises a significant concern regarding potential cancer risks attributed to radiation exposure~\cite{shah2008alara}. Sparse-view CT (SVCT) reconstruct images from a reduced number of projection views, can effectively control the radiation dose under a reasonable threshold. 
Furthermore, sparse-view CT can intriguing more possibilities in emerging fields like spectral CT, where innovative techniques such as alternating kVp switching~\cite{kim2014sparse} and dynamic beam 
blocker~\cite{lee2016moving} are employed.Moreover, in applications such as C-arm CT or dental CT, where scanning time is predominantly determined by the slower flat-panel detector, sparse-view CT emerges as a promising solution for speeding the scanning process\cite{pan2009commercial}.

\begin{figure}[H]
    \centering
    \includegraphics[width=1\linewidth]{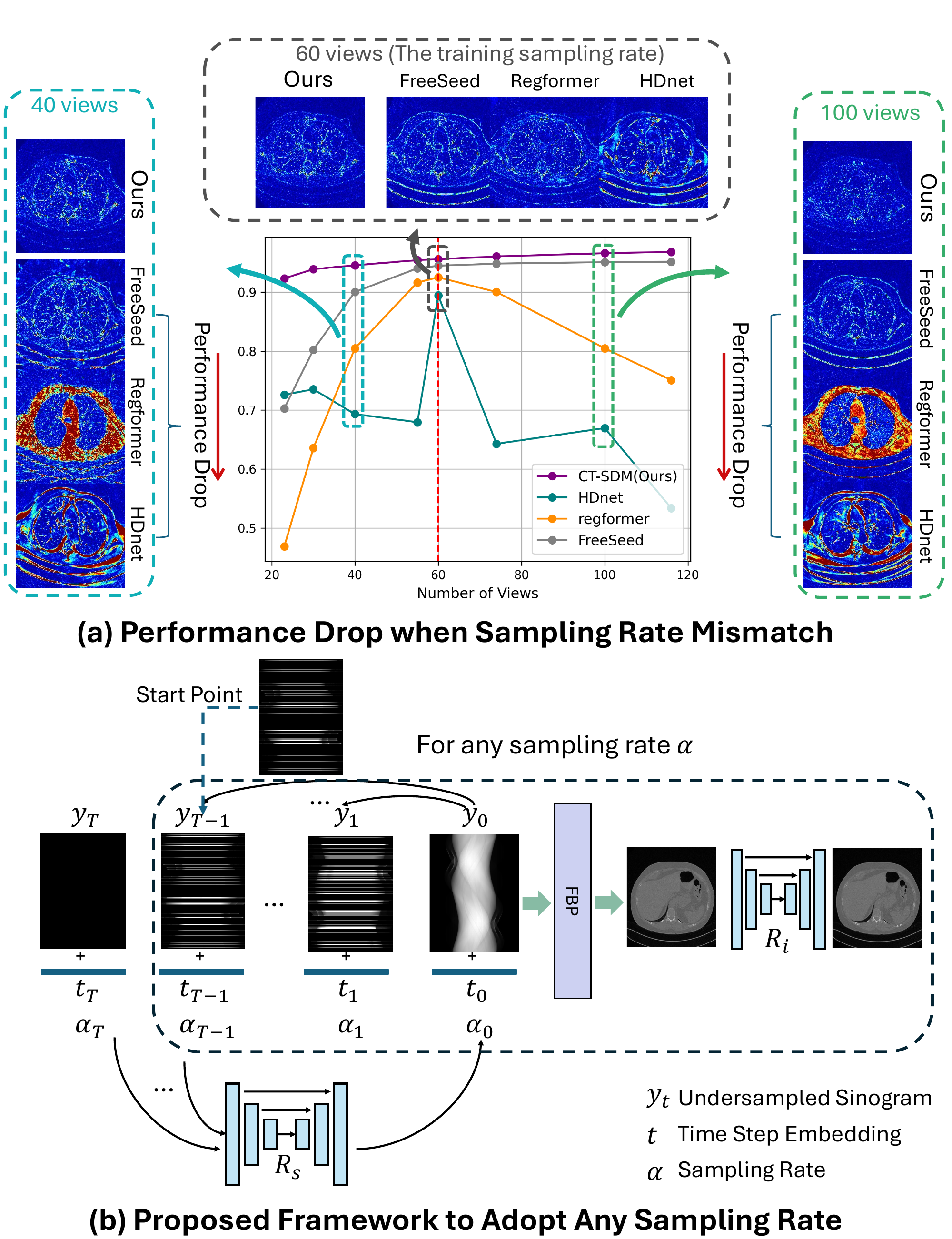}
    \caption{(a) DL based SVCT reconstructions are face performance drop when train and test on different sampling rates (i.e, Sampling Mismatch). (b) The proposed method the proposed method replace the image degradation in forward process of diffusion model as the projection view sampling. Choosing a certain sampling rate as a start point at inference, the proposed model can reconstruct SVCT images under any sampling rates.}
    \label{fig:mtivation}
\end{figure}

However, while the number of projection views is reduced during scanning, server artifacts would appear in reconstructed images which can strongly undermine clinical diagnosis. To improve the image quality of SVCT, model based methods are proposed in the early researches. Tian et al.~\cite{tian2011low} proposed EPTV (edge-preserving TV norm), which enhances edge information by incorporating adaptive weights. 
Yang et al. developed a high-order total variation (TV) minimization algorithm. Wang et al.~\cite{wang2017reweighted} introduced an iterative reweighted anisotropic total variation (RwATV) method.  As the developments of the deep learning methods in recent years, deep reconstruction methods have achieved promising performance for sparse-view CT imaging. Some of these methods~\cite{han2018framing,kang2017deep,zang2018super,zhao2018sparse} train an image model to remove artifacts from reconstructed SVCT images, while others unfold the iterative reconstruction~\cite{he2018optimizing,chen2018learn,xia2023transformer} into a deep network and learn the reconstruction parameters in end-to-end training. 
However, the prevailing trend among deep learning reconstruction methods is their training and testing within specific sampling rate configurations, often overlooking the model's generalization performance across diverse sampling rates. This oversight is exemplified in Fig~\ref{fig:mtivation}(a), where deep learning-based Sparse-View CT (SVCT) reconstructions manifest performance degradation when confronted with sampling rate mismatches, presenting a significant challenge for SVCT deployment in clinical settings. Given the paramount importance of CT imaging as a diagnostic tool, it is imperative for it to exhibit flexibility across various examination requirements. Yet, accommodating this versatility by training and storing model parameters across all requisite setups is rendered impractical due to prohibitive training times and memory overheads.

For this reason, this paper introduces a sampling-rate-adaptive method based on diffusion models, which employs a single model capable of accommodating all sampling rates. Specifically, the proposed approach substitutes the conventional degradation process in the forward step of the diffusion model (i.e., usually the Gaussian noise) with the projection view sampling. Subsequently, the inference stage (i.e., the reverse process) of the diffusion model is designed to gradually recover full-sampled data. As shown in~\ref{fig:mtivation}(b), by selecting a specific sampling rate as the starting point during inference, the proposed model can effectively reconstruct SVCT images across various sampling rates.

Moreover, challenge also lies in effectively allocating the sampling rate and specific projection views to each iteration step. This task is critical for achieving optimal reconstruction performance across various sampling rates. To address this challenge, we introduce a grouped-random sampling strategy. This strategy is designed to dynamically adjust the sampling rate and select projection views during the training process. It divides all sampling views into orderly and equally spaced groups, facilitating comprehensive and partially random angle coverage. By sequentially selecting whole groups of sampling views and resorting to random selection within groups when necessary, our strategy ensures both diversity and uniformity in views selection. This not only mitigates bias but also provides valuable data augmentation during training, promising more robust and accurate results across diverse sampling scenarios.

Our contributions can be summarised as follows:

\begin{itemize}
    \item We propose a sampling-rate-adaptive method based on diffusion models, which employs a single model capable of accommodating all sampling rates.
    \item We introduce a novel grouped-random sampling strategy for dynamically adjusting the sampling rate and selecting projection views during training, addressing biases and providing valuable data augmentation.
    \item We demonstrate the effectiveness of our approach through comprehensive experimentation, showcasing improved reconstruction performance across diverse sampling scenarios in sparse-view CT reconstruction.
\end{itemize}

\section{Related works}
\subsection{Deep learning based CT Reconstruction}
The exploration of CT reconstruction has been significantly enriched by the deployment of deep learning strategies, as evidenced in a variety of studies~\cite{jin2017deep,kang2017deep,hu2020hybrid,lee2017view,yang2022sparse,zhang2020metainv}. These approaches can be divided into two categories: image domain enhancement and dual-domain reconstruction. The former leverages denoising techniques from low-level vision to develop networks aimed at artifact reduction and image clarification. Jin \textit{et al.}~\cite{jin2017deep} proposed a CNN framework inspired by the U-net architecture, designed to refine images reconstructed via FBP (Filtered Back Projection) through the use of CNNs. Similarly, Han \textit{et al.}~\cite{han2018framing} target the restoration of high-frequency edges in sparse-view CT, employing wavelet transforms within deep learning models to effectively diminish artifacts. Chen \textit{et al.}~\cite{chen2017low} introduced a residual encoder-decoder CNN (REDCNN) tailored for low-dose scenarios. 

On the flip side, dual-domain reconstruction methods utilize data from both the sinogram and image domains to overcome the over smoothing tendencies of CNN architectures. Hu \textit{et al.}~\cite{hu2020hybrid} proposed the HDNet, a hybrid-domain network that deconstructs the SVCT reconstruction problem into simpler, sequential tasks. Yang \textit{et al.}~\cite{yang2022inner} conceived the Sinogram Inner-Structure Transformer, capitalizing on the inherent structures within the sinogram domain to mitigate noise in low-dose CT (LDCT) images. Additionally, AUTOMAP~\cite{zhu2018image} and IRadonMap~\cite{he2020radon} both present convolutional network-based solutions that facilitate the translation of measurement data directly into image representations.
\subsection{Diffusion Model in Image Reconstruction}
To date, diffusion models have been found to be highly powerful across various domains, ranging from generative modeling tasks such as image generation~\cite{dhariwal2021diffusion}, image super-resolution~\cite{li2022diffusion}, and image inpainting~\cite{lugmayr2022repaint} to discriminative tasks such as image segmentation~\cite{amit2021segdiff}, classification~\cite{zimmermann2021score}, and anomaly detection~\cite{wolleb2022diffusion}. Recently, in medical imaging, researchers have also proposed diffusion model-based methods.
Kim et al.~\cite{kim2022diffusemorph} proposed DiffuseMorph, a diffusion-based method for medical image registration. DiffuseMorph integrates the diffusion network and the deformation network into an end-to-end training framework. The diffusion network scores the deformations between the moving and fixed images, while the deformation network uses this information to estimate the deformation field.
Packhäuser et al.~\cite{packhauser2023generation} utilized a latent variable diffusion model~\cite{rombach2022high} to generate high-quality class-conditional chest X-rays, preserving the privacy of sensitive biological features while generating samples.

Meng et al.~\cite{meng2022novel} proposed a unified multi-modal conditional generation method (UMM-CSGM) to generate missing modalities in multi-modal data. This method uses all available modalities as conditions to synthesize the missing modalities, utilizing a conditional model based on SDE~\cite{song2020score} to learn cross-modal conditional distributions.
Chung et al.~\cite{chung2022score} applied score-based diffusion models to solve the inverse problem of image reconstruction in fast MRI scans. This method first uses denoising score matching to train a continuously time-varying score function on magnitude images; then, in the reverse SDE process, it employs the pre-trained score model and uses the VE-SDE method~\cite{song2020denoising}, taking undersampled data as conditional input to generate reconstructed images. 
Liu et al.~\cite{liu2023dolce} used the DDPM model in the limited-angle CT reconstruction problem and proposed the DOLCE method. This method takes the FBP reconstructed images from limited-angle sinograms as prior information to modulate the diffusion model and uses the consistency condition of the sinogram with an $l_2$ norm loss in the iterative steps of inference to ensure consistency.
\section{Methodology}
\begin{figure}[]
    \centering
    \includegraphics[width=1\linewidth]{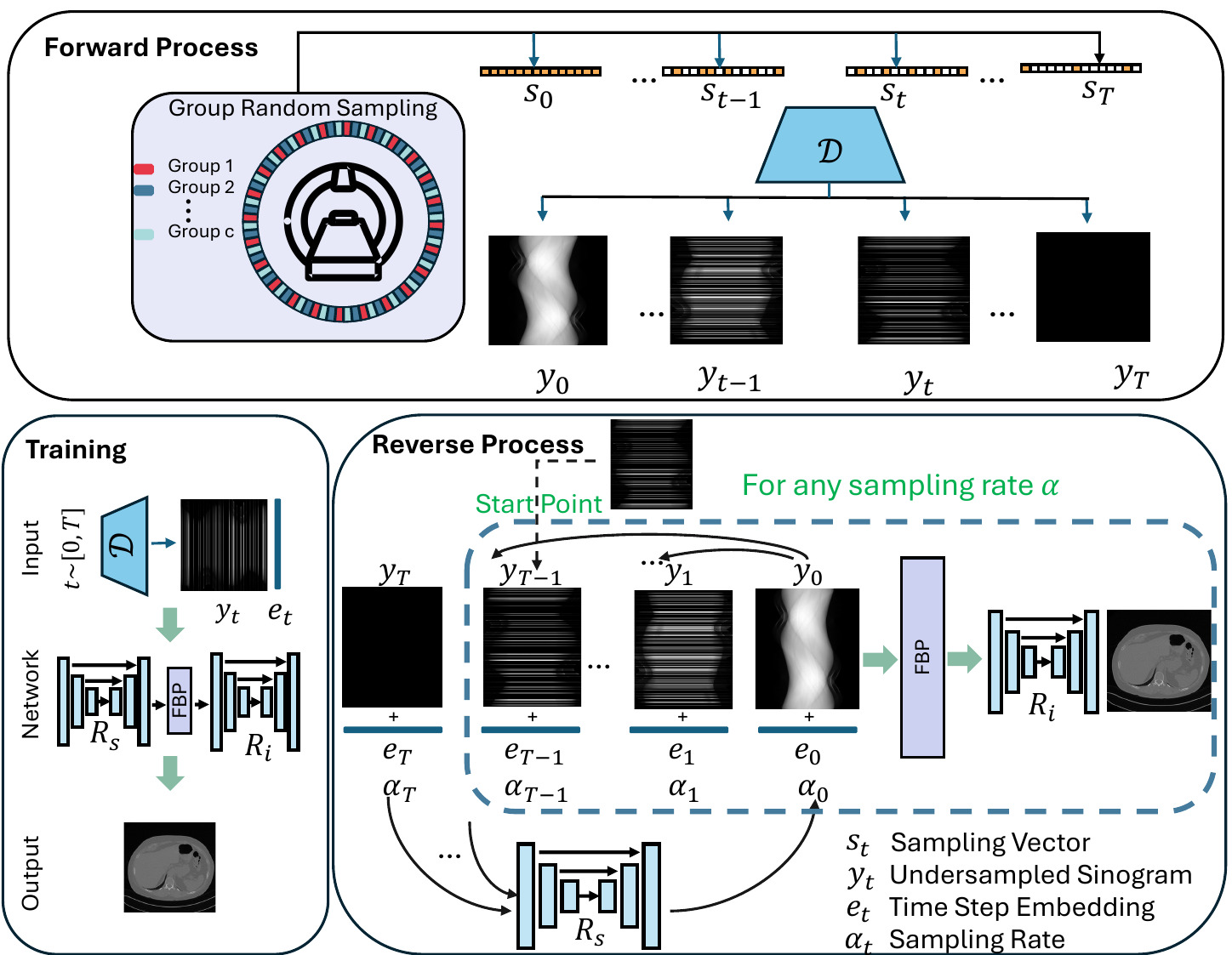}
    \caption{The overall framework of the proposed methods. The forward process of Sampling Diffusion Model is to determine the 
undersampled CT measurements (i.e., sinograms) at each sampling step $t$. And a sampling step $t$ is corresponded to a specific sampling rate $\alpha_{t}$. The reverse process is aims to recover the full-view sinogram $y_{0}$ from the measurements $y_{T}$ obtained at a certain start sampling rate $\alpha_{T}$, utilizing the networks trained for undersampled data recovering. Both the forward and reverse process diffusion are designed in sinogram domain to allow the simulation of accoutre data sampling.}
    \label{fig:framework}
\end{figure}
\subsection{Problem Formulation}
The imaging of X-ray CT can be views as a Radon transform projecting the CT images $x$ into measurements $y$ at different imaging angles:

\begin{equation}
    y = Ax + \eta,
\end{equation}
where $x\in \mathbb{R}^{n}$, $n = w\times h$, and $w$ $h$ are the width and height of $x$. $A\in \mathbb{R}^{m \times n}$ is the Radon transform matrix, where $m = v \times d$, $v$ is the number of projection views, and $d$ is the number of detectors. $\eta$ is the noise. Thus, CT reconstruction is to solve $x$ from the measurements $y$. With sufficient sampling rate (i.e., sufficient projection views $v$) and high Signal-Noise Ratio (i.e., low noise $\eta$), image $x$ can be reconstructed with good quality for clinical usage. However, more projection views cause higher radiation dose which raise the potential cancer risks. Thus,  Sparse-View CT technique is trying to reconstruct images from a undersampled measurements: 
\begin{equation}
    y_{s} = A_{s}x + \eta_{s} ,
\end{equation} 
 where $A_{s}$ represents the Radon transform matrix with a reduced number of projections $v_{s} < v$, $\eta_{s}$ is the noise under sparse view sampling. 

 \subsection{Cold Diffusion Model}

The Cold Diffusion Model (CDM) \cite{bansal2024cold} represents a significant extension of traditional diffusion models, which are fundamentally based on the interplay between degradation and restoration processes of images. At its core, the standard diffusion model employs an initial process degrades images by introducing Gaussian noise, followed by a denoising process developed through rigorous training for image restoration. This iterative application of degradation and restoration mechanisms facilitates the generation of images and achieves promising performance.

In contrast to the conventional diffusion models that limit the degradation process to the Gaussian noise, the CDM extends the degradation operator to include a wide variety of transformations such as blurring, animorphing, and masking, thus accommodating a broader spectrum of degradation effects. This approach allows for a more diverse and realistic representation of image degradation.

Consider an image \(x_{0} \in \mathbb{R}^{N}\). The degradation of \(x_{0}\) by an operator \(D\), under sampling step \(t\), is expressed as:
\begin{equation}
x_{t} = D(x_{0}, t),
\end{equation}
where \(D(x_{0}, t)\) is expected to vary continuously with \(t\), ensuring that \(D(x_{0}, 0) = x_{0}\).

To recover images from a degraded conditions, the CDM employs a restoration operator \(R\) that aims to approximately reverse the effects of \(D\). The operator \(R\) seeks to fulfill the condition:
\begin{equation}
R(x_{t}, t) \approx x_{0}.
\end{equation}
In practice, this restoration function is implemented through a neural network parameterized by \(\theta\), which is optimized to minimize the difference between the degraded and original images, represented as:
\begin{equation}
\min_{\theta} \|R_{\theta}(x_{t}, t) - x_{0}\|.
\end{equation}

This generalized framework of image degradation and restoration forms the foundation of the Cold Diffusion Model, providing a versatile and powerful tool for low-level vision tasks such as deblurring, inpainting and super-resolution.



 \subsection{Sampling Diffusion Model for CT Reconstruction}

Following the principles of the Cold Diffusion Model, we conceptualize the sampling processes of CT imaging as a category of image degradation operators, namely Sampling Diffusion Model for SVCT (CT-SDM). In CT-SDM, both the forward and reverse process are designed in sinogram domain to allow the simulation of accoutre data sampling. This approach allows us to endow the degradation operator with practical physical meanings that directly correspond to data sampling practices in CT imaging. Consequently, each step of the Sampling Diffusion Model can be analogously mapped to a specific sampling rate within the sparse-view CT imaging, thus bridging a methodological link between image degradation and X-ray projections. 
The overall architecture are shown in Fig \ref{fig:framework}. 

\subsubsection{Degradation Operator of Sampling Diffusion}
The forward process of Sampling Diffusion Model is to determine the 
undersampled CT measurements (i.e., sinograms) at each sampling step $t$. And a sampling step $t$ is corresponded to a specific sampling rate $\alpha_{t}$, which represents the number of sampled views during CT imaging.
Given the Full-sampled CT image $x$, the degradation operator of forward process can be expressed as:

\begin{equation}
\begin{split}
    y_{t} &= D(y,t) = D(A_{\alpha_{t}}x + \eta_{t},t), \qquad t= 0,1,...,T,
\end{split}
\end{equation}
where, $y_{t}$ signifies the measurements obtained under the sampling step $t$, $\alpha_{t}$ is the sampling rate corresponding to the step $t$, and $A_{\alpha_{t}} \in \mathbb{R}^{m_{\alpha_{t}} \times n}$ represents the Radon transform adapted to the sparse-view setup characterized by the reduced number of projections $v_{\alpha_{t}} < v$. The term $\eta_{t}$ denotes the noise associated with the sparse-view imaging process at the corresponding sampling rate. 

By defining the forward diffusion process in such a manner, we not only adhere to the Cold Diffusion Model's theoretical framework but also directly align it with practical imaging scenarios in CT, particularly emphasizing the importance of managing radiation exposure through the adjustment of sampling rates. 

\subsubsection{Restoration Operator of Sampling Diffusion}
The reverse process is aims to recover the full-view sinogram $y_{0}$ from the measurements $y_{t}$ obtained at a certain start sampling rate $\alpha_{t}$. Thus, a restoration operator is designed to reverse the effects of degradation, in this case, the under-sampling inherent in sparse-view CT imaging. A CNN $R_{s}$ (Detailed in Section. \ref{sec:net}) is defined as the restoration operator, which takes a undersampled sinogram  $y_{t}$ at sampling step $t$ as input, to estimate the full-view sinogram:

\begin{equation}
    \hat{y_{0}} = R_{s}(y_{t}, t),
\end{equation}
where $\hat{y_{0}}$ is a coarse estimation of full-view sinogram. In the reverse process, we take a iterative approach to generally make $\hat{y_{0}}$ to approximate $y_{0}$. As shown in Fig \ref{fig:framework}, start form a very low sampling rate $\alpha_{T}$, the restoration operator first recover a coarse estimation $\hat{y_{0}}$. 

Then, iterative reverse process is conducted in sinogram domain. Since each step of estimating $\hat{y_{0}}$ is not perfect, we use the Transformation Agnostic Cold Sampling (TACoS)~\cite{bansal2024cold,huang2023cdiffmr} during inference to avoid noise accumulation. The TACoS is shown in Algorithm \ref{5al2}. Then, the final estimated $\hat{y_{0}}$ is then reconstructed into images and refined by the image domain network $R_{i}$. 

\begin{algorithm}
\caption{\textbf{TACoS}}
\begin{algorithmic}[1]
\State \textbf{Input:} undersampled sinogram $y_t$
\For{$s = t, t-1, \ldots, 1$}
    \State $\hat{y}_0 \leftarrow R_s(y_s, s)$
    \State $\hat{y}_{s-1} = x_s - D(\hat{y}_0, s) + D(\hat{y}_0, s-1)$
\EndFor

\end{algorithmic}
\label{5al2}
\end{algorithm}



\begin{figure}
    \centering
\includegraphics[width=1\linewidth]{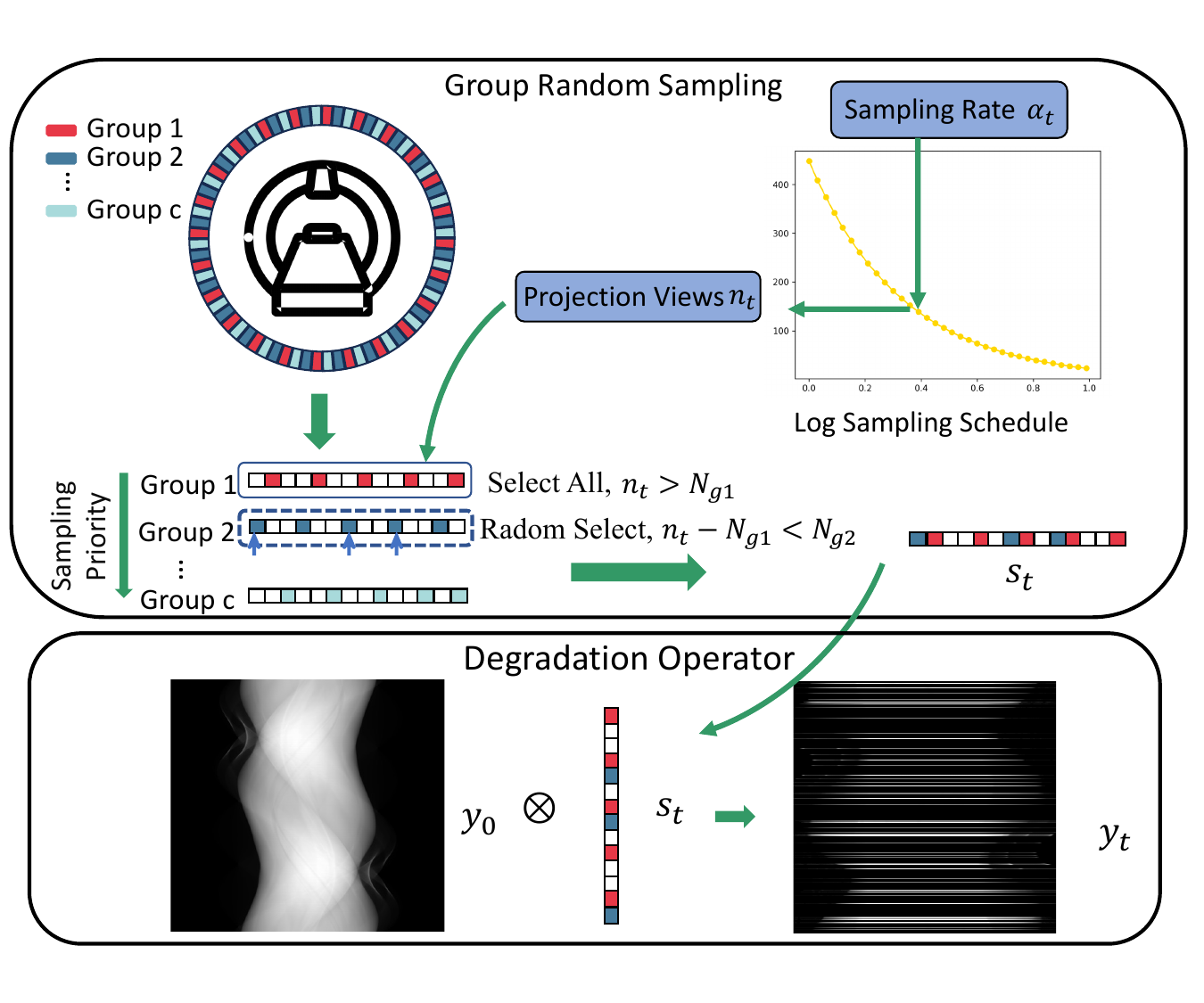}
    \caption{The Group-Random Sampling Schedule. The grouped-random sampling method divides all sampling angles into $c$ orderly and equally spaced groups to ensure comprehensive and partially random angle coverage. }
    \label{5fig:group}
\end{figure}
\subsection{Grouped-Random Sampling Strategy}
\label{group}
To effectively allocate the sampling rate $\alpha_{t}$ and specific projection views corresponding to each iteration step $t$, we 
 introduces a grouped-random sampling strategy as a dynamic training method. Initially, we gradually backtrack the reverse process by  setting the mapping of the sampling rate and iteration steps exponentially~\cite{huang2023cdiffmr} to adjust the variation in sampling rate, as shown in Figure~\ref{5fig:group}. Since the process of restoring images $\hat{y_{0}}$ at low sampling rates is relatively more complex, the exponential relationship causes a slower increase in the sampling rate at lower values and a faster increase at higher values. 

Subsequently, the grouped-random sampling method divides all sampling angles into $c$ orderly and equally spaced groups $g_{1},g_{2},...,g_{c}$. Let $Y=[y_{1},y_{2},y_{3},...,y_{v}]$, then $g_{1}=[y_{1},y_{c},y_{2c},...y_{\lfloor  \frac{v}{c} \rfloor}]$, $g_{2}=[y_{2},y_{c+1},y_{2c+1},...y_{\lfloor \frac{v}{c} \rfloor+1}]$, and so on.$ \lfloor \rfloor$ denotes the floor function, which rounds down to the nearest integer. The number of projections is determined according to the target sampling rate of the given iteration step $t$. Whole groups of sampling views are selected sequentially from $g_{1}, g_{2}, \ldots$. When the remaining unselected views are fewer than the size of a full group, views are randomly selected from within the current group. This procedure ensures comprehensive and partially random angle coverage, preventing bias from uniform or deterministic selection processes and providing data augmentation for the training process. This process can be described by the following algorithm:

\begin{algorithm}
\label{5al1}
\caption{\textbf{Grouped-Random Sampling Strategy.}}
\begin{algorithmic}[1]
\State \textbf{Input:} Number of sampling views $k$, fully sampling views $v$, number of group $c$
\For{$s = 1, 2, \ldots, c$}
 \If{$s \times\lfloor  \frac{v}{c} \rfloor  \leq  k $}
    \State Select the entire group $g_{s}$
   \Else
   \State Randomly select $k-(s-1)\times \lfloor  \frac{v}{c} \rfloor$ projection views within group $g_{s}$
   \State BREAK
    \EndIf
\EndFor
\end{algorithmic}
\end{algorithm}

Here, the number of projections $k$ is determined by the sampling rate $\alpha_{t}$, $k=\alpha_{t} \times v$.

\begin{figure*}

  \centering            
  \includegraphics[width=1\textwidth]{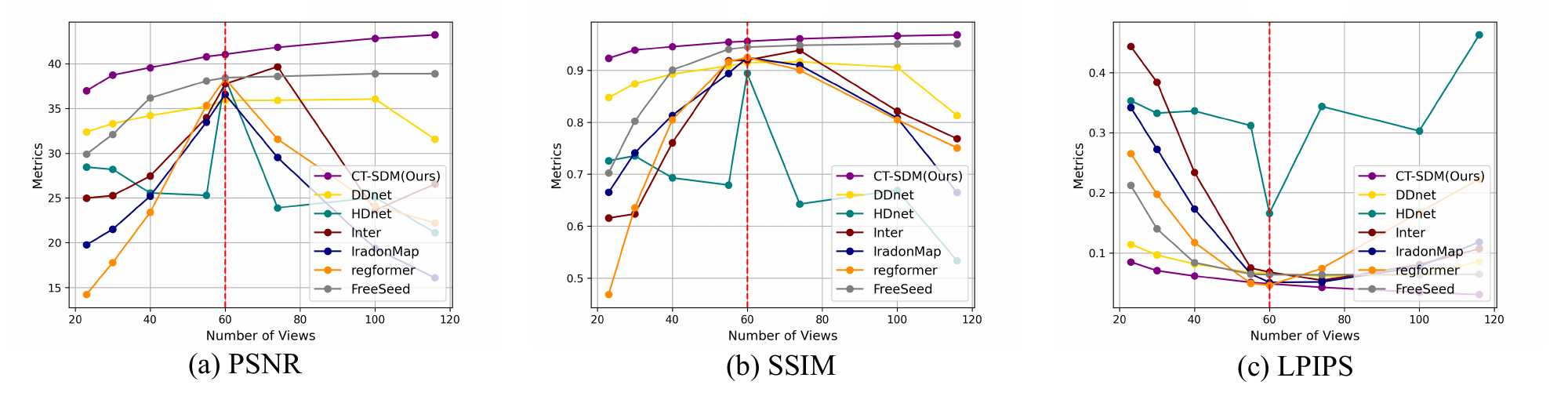}

  \caption{Performance comparison of various methods at different sampling rates on the LDCT dataset. Each method is trained under 60 views(red line), and the results are evaluated on test data with varying numbers of projections: 116, 100, 74, 60, 55, 40, 30, and 23. }    
  \label{5fig:line1}            
\end{figure*}



\begin{figure*}

 \includegraphics[width=1\textwidth]{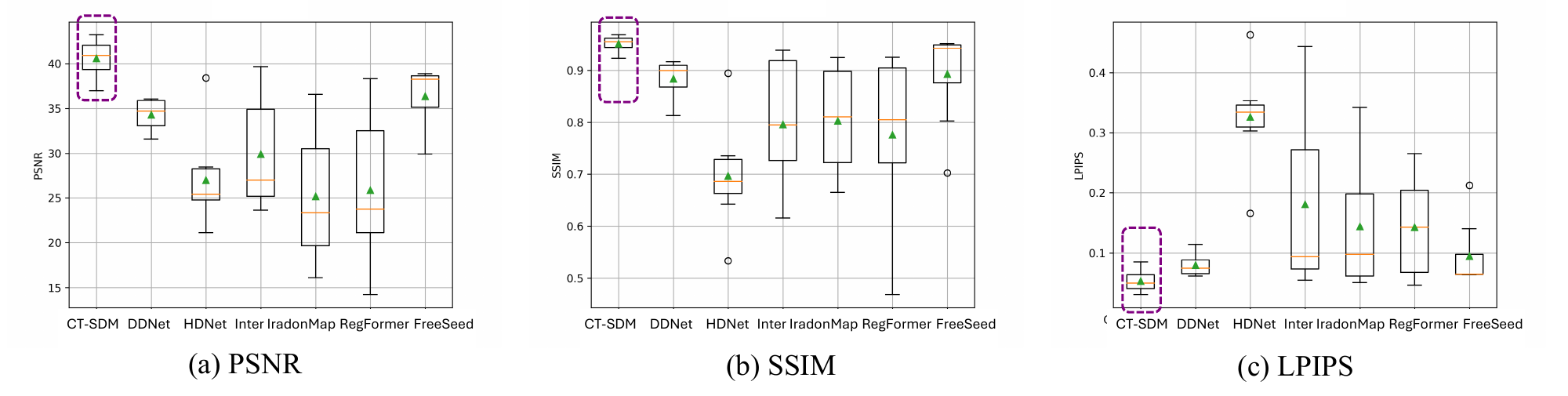}
  \caption{Box plots showing the performance of various methods at different sampling rates on the LDCT dataset. Each method is trained under 60 projections, and the results are evaluated on test data with varying numbers of projections: 116, 100, 74, 60, 55, 40, 30, and 23. }    
  \label{5fig:box_ldct}            
\end{figure*}

The main advantage of the grouped-random sampling strategy lies in its ability to simulate the variability of available projection views in real-world scenarios, which is common in clinical settings due to patient movement, equipment limitations, or specific diagnostic needs. By introducing randomness within a structured grouping framework, the model extensively simulates potential imaging conditions, thereby enhancing its adaptability and performance in various sparse-view CT configurations. Additionally, this strategy serves as a powerful form of data augmentation, expanding the diversity of training data without the need for additional physical scans. This augmentation is crucial for training deep learning models to accurately reconstruct images from sparsely sampled sinogram data, effectively improving reconstruction accuracy and reducing artifacts in the final images.

\subsection{Network Training}
\label{sec:net}
The proposed method contains two CNN to handle the SVCT reconstruction task in both the sinogram domain and the image domain. The network in sinogram domain is the restoration operator $R_{s}$ to estimate the full-view sinogram $\hat{y_{0}}$ from the input $y_{t}$. Thus, the network parameters $s$ are optimized by minimizing the loss function:
\begin{equation}
\begin{split}
   L_{s}& = \Vert R_{s}(y_{t})-y\Vert\\,
&     = \Vert R_{s}(D(y,t),t)-y\Vert, \qquad t= 1,2,...,T.
\end{split}
\end{equation}

The estimated full-view sinogram $\hat{y_{0}}$ is then reconstructed into images and a image domain network $R_{i}$ is used for image domain refinements. The loss function can be defined as follows:
\begin{equation}
L_{i} = \Vert R_{i}(A^{\dagger}\hat{y_{0}})-x \Vert,
\end{equation}
where $A^{\dagger}$ is the pseudo-inversion of Radon transform $A$, which can be implemented by Filtered Back-Projection (FBP)~\cite{xia2023transformer}. 
These two networks are jointly trained together:

\begin{equation}
L = L_{s}+\lambda L_{i},
\end{equation}
where $\lambda$ is the control the ratio of two parts.

\section{Experiments}
\subsection{Datasets}
To validate the performance of the proposed methods,  \textit{Low-Dose CT Image and Projection Datasets} (LDCT Datasets)~\cite{moen2021low} and the \textit{TCGA-KIRC}~\cite{clark2013cancer,Akin2016a} dataset are used.

The LDCT Dataset consists of CT scans from three common exam types: non-contrast head CT scans acquired for acute cognitive or motor deficit, low-dose non-contrast chest scans acquired to screen high-risk patients for pulmonary nodules, and contrast-enhanced CT scans of the abdomen acquired to look for metastatic liver lesions. It contains 25,141 CT images in total from 150 patients. 120 patients are randomly selected into the training set and remains are for testing.

The TCGA-KIRC dataset consists both CT and MRI of scans for Kidney Renal Clear Cell Carcinoma. In our experiments, we use 15153 CT images from 50 participants to generate sinograms with the same configuration as LDCT Datasets. 40 patients are randomly split into the training set and remains are for testing. 

\subsection{Implementation Details}
 In this paper, we use an attention-based UNet with time embedding module~\cite{song2020denoising} for the restoration operator. In image domain, we use one ResNet Block~\cite{he2016deep} implemented in original CDM~\cite{bansal2024cold} for a light weight image refinements. As radiation dose of each view is not reduced in our experimental settings, the noise $\eta_{s}$ is small and we ignore it for simplification. During training, $\lambda$ is set to 1.

The Torch-Radon~\cite{torch_radon} toolbox is used for sinogram generation. We use a fan-beam CT geometry with 488 projection views and 736 detectors as the full view sampling setup. The source-to-detector distance was set to 1000 mm, and the source-to-rotation-centre distance was 512 mm. The reconstructed image resolution is set to $256 \times 256$ pixels.

\subsection{Evaluation Metrics}
For the quantitative evaluation, we employ the Peak Signal-to-Noise Ratio (PSNR), Structural Similarity Index Measure (SSIM), and Learned Perceptual Image Patch Similarity (LPIPS)~\cite{Zhang2018LPIPS} as metrics to gauge the quality of reconstruction. It's worth mentioning that LPIPS, with its foundation in deep learning-based perceptual metrics, mirrors human visual perception closely, offering a detailed assessment of the fidelity in image reconstruction.
\subsection{Performance under different sampling rates}
To validate the generalization ability of the proposed method on data with different sampling rates, this section compares six methods: DDNet~\cite{zhang2018sparse}, View-Inter~\cite{lee2017view}, HDNet~\cite{hu2020hybrid}, IradonMap~\cite{he2020radon}, RegFormer~\cite{xia2023transformer}, and FreeSeed~\cite{ma2023freeseed}. 

DDNet~\cite{zhang2018sparse} is a representative image-domain method aimed at removing artifacts and improving image quality from FBP reconstructed images. View-Inter~\cite{lee2017view} is a sinogram-domain method that completes the full projection angle sinogram only in the sinogram domain. HDNet~\cite{hu2020hybrid} is a typical dual-domain method that interpolates sparse view sinogram data and uses two CNNs to refine image quality. IradonMap~\cite{he2020radon} is essentially a learnable Iradon transform combined with an image-domain CNN. RegFormer~\cite{xia2023transformer} is a deep unfolding reconstruction method combined with Swin-Transformer~\cite{liu2021swin}. FreeSeed~\cite{ma2023freeseed} is an image-domain post-processing method that combines frequency-domain adaptive bandpass filtering with an image-domain reconstruction network.

During the training of the comparative methods, the number of projection is set to 60 (i.e., sampling rate of $60/448=13.4\%$) was selected to generate the corresponding sparse-view sinograms for network training. The proposed method, using SDM to gradually add projection angles, does not require setting a sampling rate during training. Instead, it can achieve reconstruction at the corresponding sampling rate during testing by selecting the diffusion starting point corresponding to the desired sampling rate.

During testing, the performance of the methods was evaluated on test images by generating sinograms with projection numbers of 116, 100, 74, 60, 55, 40, 30, and 23, respectively, to test their performance at different sampling rates. 
\subsubsection{Experimental Results on the LDCT Dataset}

\begin{table*}[!h]
\begin{center}
\linespread{1.5}
\caption{Mean and Standard Deviation of Reconstruction Performance of Different Methods at Various Sampling Rates on the LDCT Dataset and TCGA-KIRC Dataset.The best results are \textbf{highlighted}, and the second results are \underline{underlined}. $\downarrow$ ($\uparrow$) indicates that lower (higher) is better. * denotes results that are significantly different from the best results by the Wilcoxon test ($p<0.05$). }
\label{5table:ldct&kirc}
\begin{tabular}{ccccccc}
\hline
&\multicolumn{3}{|c}{LDCT(Averaged over Sampling Rates)}&\multicolumn{3}{|c}{TCGA-KIRC(Averaged over Sampling Rates)}\\
\hline
 Methods & \multicolumn{1}{|c}{PSNR($\uparrow$)}& SSIM($\uparrow$) & LPIPS($\downarrow$)& \multicolumn{1}{|c}{PSNR($\uparrow$)} & SSIM($\uparrow$) & LPIPS($\downarrow$) \\
\hline
DDnet* & 34.314\scriptsize{$\pm$}1.630 & 0.884\scriptsize{$\pm$}0.035 & 0.079\scriptsize{$\pm$}0.018&  \underline{36.783\scriptsize{$\pm$}2.321} & \underline{0.901\scriptsize{$\pm$}0.041} & \underline{0.073\scriptsize{$\pm$}0.013} \\

Inter* & 29.909\scriptsize{$\pm$}5.864 & 0.796\scriptsize{$\pm$}0.120 & 0.181\scriptsize{$\pm$}0.145& 30.265\scriptsize{$\pm$}5.682 & 0.832\scriptsize{$\pm$}0.106 & 0.161\scriptsize{$\pm$}0.120 \\

HDnet* & 27.007\scriptsize{$\pm$}4.834 & 0.697\scriptsize{$\pm$}0.095 & 0.326\scriptsize{$\pm$}0.076& 28.776\scriptsize{$\pm$}5.179 & 0.783\scriptsize{$\pm$}0.079 & 0.268\scriptsize{$\pm$}0.069 \\
IradonMap* & 25.192\scriptsize{$\pm$}6.862 & 0.803\scriptsize{$\pm$}0.098 & 0.144\scriptsize{$\pm$}0.103& 25.611\scriptsize{$\pm$}7.983 & 0.800\scriptsize{$\pm$}0.098 & 0.149\scriptsize{$\pm$}0.106 \\
RegFormer* & 25.875\scriptsize{$\pm$}7.910 & 0.776\scriptsize{$\pm$}0.147 & 0.143\scriptsize{$\pm$}0.078 & 26.062\scriptsize{$\pm$}7.750 & 0.798\scriptsize{$\pm$}0.128 & 0.147\scriptsize{$\pm$}0.084  \\
FreeSeed* & \underline{36.376\scriptsize{$\pm$}3.253} & \underline{0.893\scriptsize{$\pm$}0.086} & \underline{0.095\scriptsize{$\pm$}0.051}& 35.718\scriptsize{$\pm$}4.162 & 0.894\scriptsize{$\pm$}0.080 & 0.095\scriptsize{$\pm$}0.059 \\

CT-SDM(Ours) & \textbf{40.612\scriptsize{$\pm$}1.978} & \textbf{0.952\scriptsize{$\pm$}0.014} & \textbf{0.053\scriptsize{$\pm$}0.017}& \textbf{40.805\scriptsize{$\pm$}1.526} & \textbf{0.966\scriptsize{$\pm$}0.011} & \textbf{0.051\scriptsize{$\pm$}0.016} \\
\hline
\end{tabular}
\end{center}
\end{table*}

he experimental results are shown in Figure~\ref{5fig:line1}. Through these experiments, we observed that while deep learning methods often exhibit excellent performance at the sampling rates set during training, their limited network generalization capability makes it difficult to maintain performance at non-training sampling rates. Particularly for dual-domain methods like RegFormer and HDNet, these algorithms perform excellently at the training sampling rate. However, errors in the sinogram domain tend to accumulate during the image reconstruction process, leading to a sharp decline in performance once the sampling rate deviates from the training setting. In contrast, pure image-domain methods exhibit better generalization ability, but their performance still noticeably degrades in experiments. Notably, increasing the sampling rate sometimes resulted in a decline in reconstruction quality, further highlighting the significant limitations of deep learning methods in terms of generalization.

The proposed SDM reduces the reliance on model parameters trained at specific sampling rates, significantly enhancing the model's generalization ability. This improvement enables the model to meet the image reconstruction requirements at different sampling rates, greatly expanding its applicability and practicality.

As shown in Figure~\ref{5fig:box_ldct}, box plots of three key performance metrics are presented for various methods at different sampling rates. The figure clearly demonstrates that the proposed method not only maintains high-quality reconstructed images under varying sampling rates but also exhibits excellent stability. In contrast, other methods like HDNet and Inter, although achieving high peak performance at specific sampling rates, exhibit large performance fluctuations between different sampling rates, indicating a lack of stability. Methods like FreeSeed, while showing better stability, still fall short of the overall performance achieved by the proposed method.

\begin{figure*}

  \centering            
  \includegraphics[width=1\textwidth]{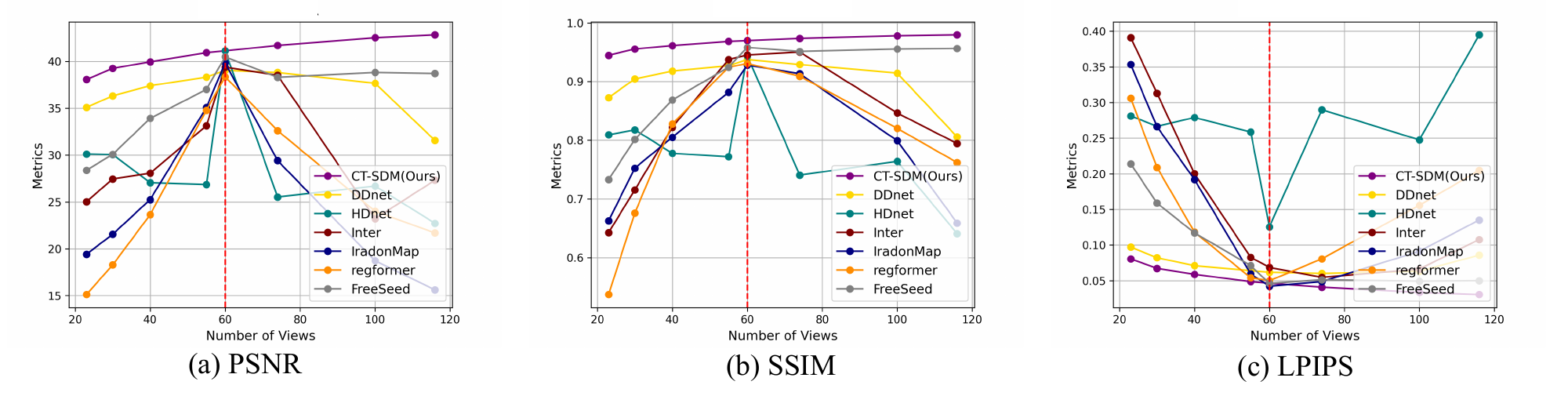}
  \caption{Performance comparison of various methods at different sampling rates on the TCGA-KIRC dataset. Each method is trained with 60 projections, and the results are evaluated on test data with varying numbers of projections: 116, 100, 74, 60, 55, 40, 30, and 23. This analysis assesses the models' ability to generalize across multiple sampling rates.}    
  \label{5fig:line2}            
\end{figure*}


\begin{figure*}[htbp]    %
  \centering            %
  
      \includegraphics[width=1\textwidth]{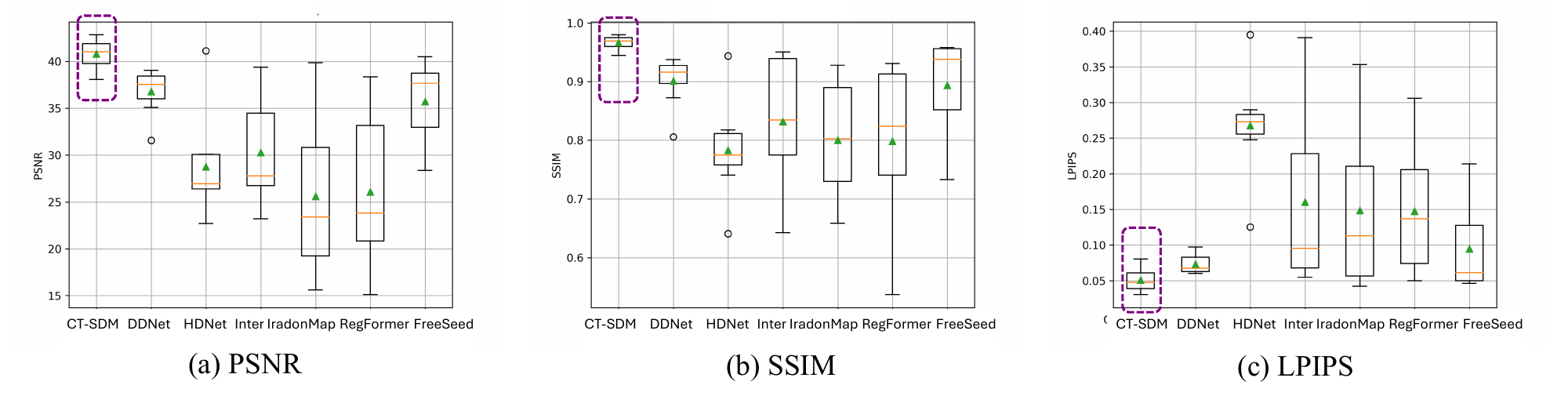}
      
  \caption{Box plots showing the performance of various methods at different sampling rates on the TCGA-KIRC dataset. Each method is trained with 60 projections, and the results are evaluated on test data with varying numbers of projections: 116, 100, 74, 60, 55, 40, 30, and 23. }    
  \label{5fig:box_kirc}            
\end{figure*}

\subsubsection{Experimental Results on the TCGA-KIRC Dataset}

The experimental results on the TCGA-KIRC dataset are shown in Figure~\ref{5fig:line2}. The performance trends of each method with varying sampling rates are generally consistent with those observed on the LDCT dataset. Dual-domain methods exhibit significant performance degradation regardless of whether the sampling rate decreases or increases, while single-domain methods experience more severe degradation at low sampling rates. Figure~\ref{5fig:box_kirc} shows the box plots of performance metrics for various methods at different sampling rates on the TCGA-KIRC dataset. With The proposed method still demonstrates superior stability and higher average performance compared to other methods.
\subsection{Performance at the Training Sampling Rate}

\begin{figure*}   
  \centering           
    \includegraphics[width=0.95\textwidth]{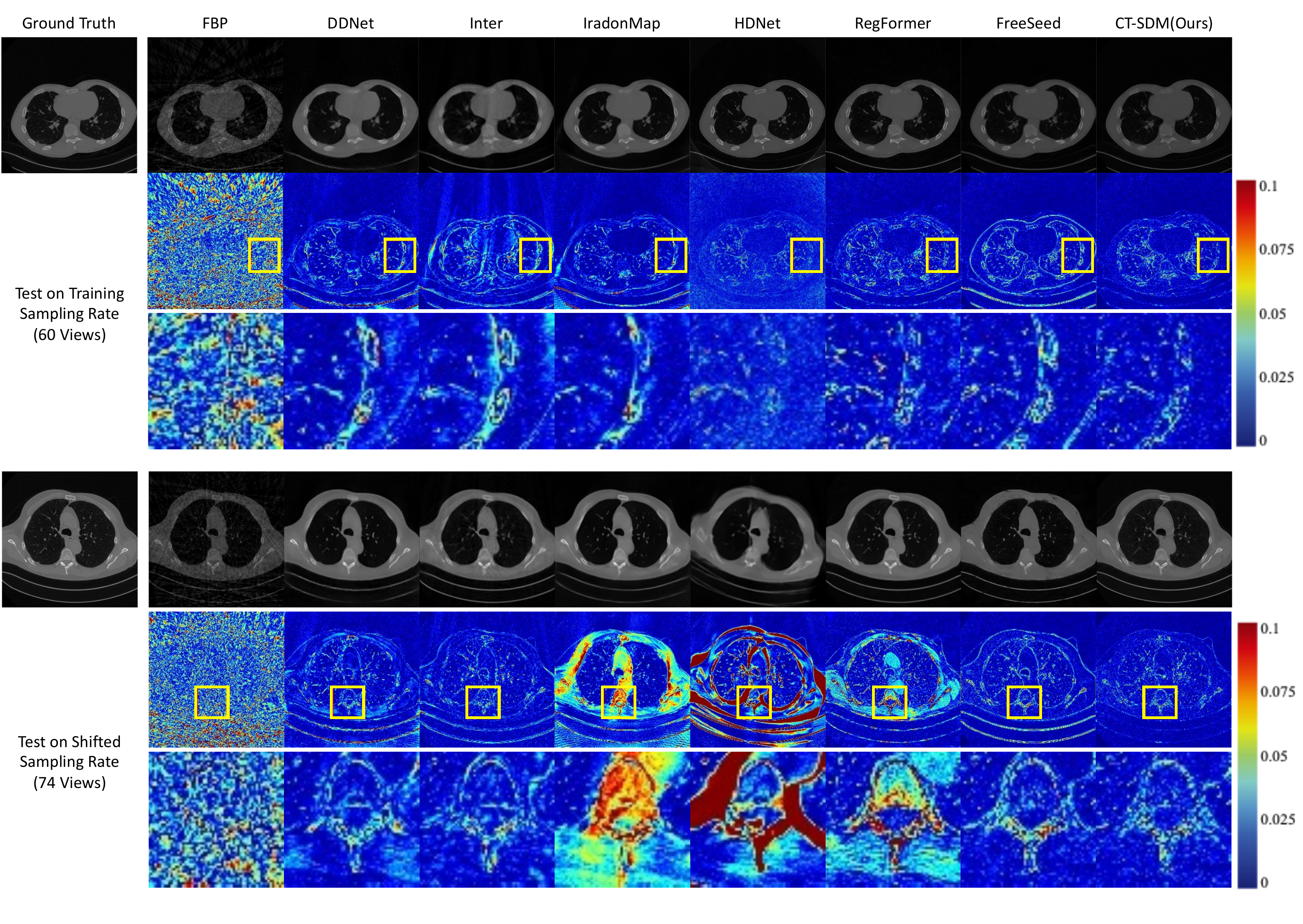}
  \caption{Visualization of reconstruction results on the LDCT dataset. We compare the reconstruction images at the training sampling rate (60 views) and the shifted sampling rate (74 views). All images are normalized and displayed.}    
  \label{5fig:compare_LDCTv2}           
\end{figure*}

  
  

Table~\ref{5table:ldct&kirc_60_view} presents the reconstruction results of various methods tested at the model's set sampling rate (60 projections), corresponding to the red dashed lines in Figures~\ref{5fig:line1} and \ref{5fig:line2}. Overall, dual-domain methods (HDNet, IradonMap, RegFormer) exhibit better reconstruction performance compared to single-domain methods (image-domain DDNet, projection-domain Inter). FreeSeed, which can be considered a post-processing method since it does not use original sinogram data, achieves good performance by performing joint processing in the frequency domain. The proposed method not only ensures that a single model performs well across all sampling rates, but also shows no significant performance loss at a single sampling rate, achieving the coparable performance across multiple metrics.

Figures~\ref{5fig:compare_LDCTv2} present visualized results and error maps on the LDCT dataset. We first compare the reconstructed images at the same sampling rate used for training. Although our method does not specify a particular sampling rate during training, it still achieves better or comparable performance to state-of-the-art methods that train and test at a fixed sampling rate. When the sampling rates are varied, all comparison methods exhibit a significant performance drop, especially the dual-domain methods (i.e., IradonMap, HDNet, and RegFormer). In contrast, the proposed method maintains good performance, demonstrating robustness and stability across different sampling rates.

\begin{table*}[!h]
\begin{center}
\linespread{1.5}
\caption{Quantitative Results of Different Methods at a Single Sampling Rate (60 Projection Views) on the LDCT Dataset and TCGA-KIRC Dataset. The best results are \textbf{highlighted}, and the second results are \underline{underlined}. $\downarrow$ ($\uparrow$) indicates that lower (higher) is better. * denotes results that are significantly different from the best results by the Wilcoxon test ($p<0.05$). }
\label{5table:ldct&kirc_60_view}
\begin{tabular}{ccccccc}
\hline
&\multicolumn{3}{|c}{LDCT (60 Projection Views)}&\multicolumn{3}{|c}{TCGA-KIRC (60 Projection Views)}\\
\hline
 Methods & \multicolumn{1}{|c}{PSNR($\uparrow$)}& SSIM($\uparrow$) & LPIPS($\downarrow$)& \multicolumn{1}{|c}{PSNR($\uparrow$)} & SSIM($\uparrow$) & LPIPS($\downarrow$) \\
\hline
 DDNet& 35.890\scriptsize{$\pm$}1.634* & 0.915\scriptsize{$\pm$}0.018* & 0.066\scriptsize{$\pm$}0.018*& 39.046\scriptsize{$\pm$}1.876* & 0.938\scriptsize{$\pm$}0.014* & 0.062\scriptsize{$\pm$}0.021* \\
 Inter & 37.746\scriptsize{$\pm$}1.698* & 0.920\scriptsize{$\pm$}0.020* & 0.068\scriptsize{$\pm$}0.021*& 39.385\scriptsize{$\pm$}1.722* & 0.945\scriptsize{$\pm$}0.015* & 0.069\scriptsize{$\pm$}0.022* \\
 HDNet & 38.426\scriptsize{$\pm$}3.313* & 0.894\scriptsize{$\pm$}0.062* & 0.166\scriptsize{$\pm$}0.088*& 40.147\scriptsize{$\pm$}2.484* & 0.944\scriptsize{$\pm$}0.041* & 0.126\scriptsize{$\pm$}0.053*  \\
  IradonMap & 36.587\scriptsize{$\pm$}1.582* & 0.925\scriptsize{$\pm$}0.017* & 0.051\scriptsize{$\pm$}0.014*& 39.866\scriptsize{$\pm$}1.766* & 0.928\scriptsize{$\pm$}0.015* & 0.049\scriptsize{$\pm$}0.019*  \\
  RegFormer & 38.316\scriptsize{$\pm$}2.050* & 0.925\scriptsize{$\pm$}0.024* & \textbf{0.046\scriptsize{$\pm$}0.020}& 38.374\scriptsize{$\pm$}1.641* & 0.931\scriptsize{$\pm$}0.021* & 0.050\scriptsize{$\pm$}0.019* \\
  FreeSeed & \underline{38.439\scriptsize{$\pm$}2.228}* & \underline{0.945\scriptsize{$\pm$}0.024}* & 0.064\scriptsize{$\pm$}0.027* & \underline{40.494\scriptsize{$\pm$}1.969}* & \underline{0.958\scriptsize{$\pm$}0.013}* & \textbf{0.046\scriptsize{$\pm$}0.014}\\
  CT-SDM (Ours) & \textbf{41.032\scriptsize{$\pm$}2.907} & \textbf{0.956\scriptsize{$\pm$}0.023} & \underline{0.049\scriptsize{$\pm$}0.023} & \textbf{41.130\scriptsize{$\pm$}1.769} & \textbf{0.970\scriptsize{$\pm$}0.015} & \underline{0.047\scriptsize{$\pm$}0.020} \\
\hline
\end{tabular}
\end{center}
\end{table*}

\begin{table*}[]
    \centering
    \caption{Comparison of different sampling strategies on 60 projection views. The best results are \textbf{highlighted}. $\downarrow$ ($\uparrow$) indicates that lower (higher) values are better. * indicates a statistically significant improvement ($p<0.05$).}
    \begin{tabular}{c|ccc}
    \hline
       Method & PSNR($\uparrow$) & SSIM($\uparrow$) & LPIPS($\downarrow$) \\ 
       \hline
        Fixed Sampling* & 40.412\scriptsize{$\pm$}2.332 & 
        0.921\scriptsize{$\pm$}0.021 & 0.053\scriptsize{$\pm$}0.021\\
        Random Sampling* & 38.012\scriptsize{$\pm$}2.667 & 0.901\scriptsize{$\pm$}0.056 & 0.066\scriptsize{$\pm$}0.043 \\
        Grouped-Random Sampling  & \textbf{41.032\scriptsize{$\pm$}2.907} & \textbf{0.956\scriptsize{$\pm$}0.023} & \textbf{0.049\scriptsize{$\pm$}0.023} \\
        \hline
    \end{tabular}
    \label{tab:group_random}
\end{table*}

\begin{figure*}
    \centering           
    \includegraphics[width=0.95\textwidth]{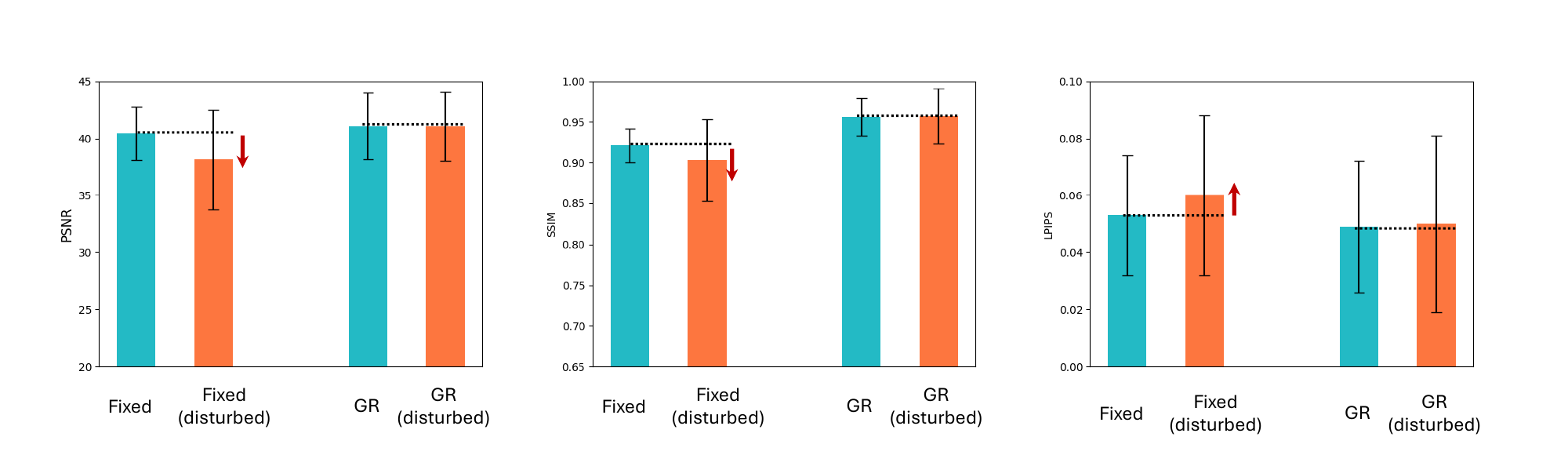}
    \caption{Robustness comparison over sampling views disturbance. Models trained with a fixed sampling strategy show a noticeable performance drop, while the proposed Grouped-Random Sampling (GR) strategy demonstrates much more stable performance.}    
    \label{dist}     
\end{figure*}
\subsection{Ablation Study}
To verify the effectiveness of the proposed method, we conduct several ablation studies in this section.

\subsubsection{Effectiveness of Grouped-Random Sampling}
As illustrated in Fig~\ref{group}, the grouped-random sampling strategy serves as a powerful form of data augmentation to improve reconstruction performance. To validate its effectiveness, we compare the performance of three sampling strategies: fixed sampling (i.e., projection views at each sampling rate are fixed and uniformly aligned), random sampling (projection views are randomly drawn from a uniform distribution), and the proposed Grouped Random Sampling. The results are shown in Table~\ref{tab:group_random}. The naive random sampling disrupts the uniformity of sampling views, undermining reconstruction performance. In contrast, the grouped-random sampling strategy introduces randomness while maintaining uniformity, thereby significantly enhancing performance compared to fixed sampling.

\subsubsection{Robustness over Sampling Views Disturbance}
The main advantage of the grouped-random sampling strategy lies in its ability to simulate the variability of available projection views in real-world scenarios, which is common in clinical settings due to patient movement, equipment limitations, or specific diagnostic needs. To verify the robustness of the proposed method, we conduct experiments using the grouped-random sampling strategy to generate the testing dataset, simulating sampling view disturbances.
The performance of fixed sampling and Grouped Random Sampling strategies is compared on view-disturbed testing data, as shown in Fig~\ref{dist}. The results demonstrate that the grouped-random Sampling strategy exhibits superior robustness to sampling view disturbances compared to fixed sampling.

\section{Conclusion}
This paper proposes a sparse-view CT reconstruction method based on  Diffusion Models, aimed at achieving high-performance reconstruction of sparse-view CT using deep learning networks at arbitrary sampling rates. While deep learning models have shown promising performance in artifact suppression for sparse-view CT, their generalization limitations make it challenging for a model trained at one sampling rate to perform well at other sampling rates. Consequently, existing methods often require models to be trained for specific sampling rates, limiting their flexibility and usability in clinical environments. To address this issue, this paper introduces the CT-SDM and employs a stepwise inference approach, enabling the same model to perform reconstructions at any sampling rate. In CT-SDM, a projection degradation operator is used to simulate the projection process in the sinogram domain, allowing the model to gradually add new projection data to highly undersampled projection data, thereby reconstructing a complete projection sinogram. By selecting a diffusion inference starting point corresponding to a specific sampling rate, our model can recover a complete projection sinogram from any sampling rate using a single trained model.Through comparative experiments on multiple public datasets, we validate that our method can achieve high-quality reconstructions of SVCT at different sampling rates, providing a flexible and reliable reconstruction solution for sparse-view imaging.

\bibliographystyle{IEEEtran}
\bibliography{sample-base}

\begin{thebibliography}{10}
\providecommand{\url}[1]{#1}
\csname url@samestyle\endcsname
\providecommand{\newblock}{\relax}
\providecommand{\bibinfo}[2]{#2}
\providecommand{\BIBentrySTDinterwordspacing}{\spaceskip=0pt\relax}
\providecommand{\BIBentryALTinterwordstretchfactor}{4}
\providecommand{\BIBentryALTinterwordspacing}{\spaceskip=\fontdimen2\font plus
\BIBentryALTinterwordstretchfactor\fontdimen3\font minus \fontdimen4\font\relax}
\providecommand{\BIBforeignlanguage}[2]{{%
\expandafter\ifx\csname l@#1\endcsname\relax
\typeout{** WARNING: IEEEtran.bst: No hyphenation pattern has been}%
\typeout{** loaded for the language `#1'. Using the pattern for}%
\typeout{** the default language instead.}%
\else
\language=\csname l@#1\endcsname
\fi
#2}}
\providecommand{\BIBdecl}{\relax}
\BIBdecl

\bibitem{shah2008alara}
N.~B. Shah and S.~L. Platt, ``Alara: is there a cause for alarm? reducing radiation risks from computed tomography scanning in children,'' \emph{Current opinion in pediatrics}, vol.~20, no.~3, pp. 243--247, 2008.

\bibitem{kim2014sparse}
K.~Kim, J.~C. Ye, W.~Worstell, J.~Ouyang, Y.~Rakvongthai, G.~El~Fakhri, and Q.~Li, ``Sparse-view spectral {CT} reconstruction using spectral patch-based low-rank penalty,'' \emph{IEEE Transactions on Medical Imaging}, vol.~34, no.~3, pp. 748--760, 2014.

\bibitem{lee2016moving}
T.~Lee, C.~Lee, J.~Baek, and S.~Cho, ``Moving beam-blocker-based low-dose cone-beam {CT},'' \emph{IEEE Transactions on Nuclear Science}, vol.~63, no.~5, pp. 2540--2549, 2016.

\bibitem{pan2009commercial}
X.~Pan, E.~Y. Sidky, and M.~Vannier, ``Why do commercial {CT} scanners still employ traditional, filtered back-projection for image reconstruction?'' \emph{Inverse Problems}, vol.~25, no.~12, p. 123009, 2009.

\bibitem{tian2011low}
Z.~Tian, X.~Jia, K.~Yuan, T.~Pan, and S.~B. Jiang, ``Low-dose {CT} reconstruction via edge-preserving total variation regularization,'' \emph{Physics in Medicine \& Biology}, vol.~56, no.~18, p. 5949, 2011.

\bibitem{wang2017reweighted}
T.~Wang, K.~Nakamoto, H.~Zhang, and H.~Liu, ``Reweighted anisotropic total variation minimization for limited-angle {CT} reconstruction,'' \emph{IEEE Transactions on Nuclear Science}, vol.~64, no.~10, pp. 2742--2760, 2017.

\bibitem{han2018framing}
Y.~Han and J.~C. Ye, ``Framing u-net via deep convolutional framelets: Application to sparse-view ct,'' \emph{IEEE Transactions on Medical Imaging}, vol.~37, no.~6, pp. 1418--1429, 2018.

\bibitem{kang2017deep}
E.~Kang, J.~Min, and J.~C. Ye, ``A deep convolutional neural network using directional wavelets for low-dose x-ray {CT} reconstruction,'' \emph{Medical Physics}, vol.~44, no.~10, pp. e360--e375, 2017.

\bibitem{zang2018super}
G.~Zang, M.~Aly, R.~Idoughi, P.~Wonka, and W.~Heidrich, ``Super-resolution and sparse view {CT} reconstruction,'' in \emph{Proceedings of the European Conference on Computer Vision (ECCV)}, 2018, pp. 137--153.

\bibitem{zhao2018sparse}
Z.~Zhao, Y.~Sun, and P.~Cong, ``Sparse-view {CT} reconstruction via generative adversarial networks,'' in \emph{2018 IEEE Nuclear Science Symposium and Medical Imaging Conference Proceedings (NSS/MIC)}, 2018, pp. 1--5.

\bibitem{he2018optimizing}
J.~He, Y.~Yang, Y.~Wang, D.~Zeng, Z.~Bian, H.~Zhang, J.~Sun, Z.~Xu, and J.~Ma, ``Optimizing a parameterized plug-and-play admm for iterative low-dose {CT} reconstruction,'' \emph{IEEE Transactions on Medical Imaging}, vol.~38, no.~2, pp. 371--382, 2018.

\bibitem{chen2018learn}
H.~Chen, Y.~Zhang, Y.~Chen, J.~Zhang, W.~Zhang, H.~Sun, Y.~Lv, P.~Liao, J.~Zhou, and G.~Wang, ``Learn: Learned experts’ assessment-based reconstruction network for sparse-data ct,'' \emph{IEEE Transactions on Medical Imaging}, vol.~37, no.~6, pp. 1333--1347, 2018.

\bibitem{xia2023transformer}
W.~Xia, Z.~Yang, Z.~Lu, Z.~Wang, and Y.~Zhang, ``Regformer: A local-nonlocal regularization-based model for sparse-view {CT} reconstruction,'' \emph{IEEE Transactions on Radiation and Plasma Medical Sciences}, 2023.

\bibitem{jin2017deep}
K.~H. Jin, M.~T. McCann, E.~Froustey, and M.~Unser, ``Deep convolutional neural network for inverse problems in imaging,'' \emph{IEEE Transactions on Image Processing}, vol.~26, no.~9, pp. 4509--4522, 2017.

\bibitem{hu2020hybrid}
D.~Hu, J.~Liu, T.~Lv, Q.~Zhao, Y.~Zhang, G.~Quan, J.~Feng, Y.~Chen, and L.~Luo, ``Hybrid-domain neural network processing for sparse-view {CT} reconstruction,'' \emph{IEEE Transactions on Radiation and Plasma Medical Sciences}, vol.~5, no.~1, pp. 88--98, 2021.

\bibitem{lee2017view}
H.~Lee, J.~Lee, and S.~Cho, ``{View-Interpolation of Sparsely Sampled Sinogram using Convolutional Neural Network},'' in \emph{Medical Imaging 2017: Image Processing}, International Society for Optics and Photonics.\hskip 1em plus 0.5em minus 0.4em\relax SPIE, 2017, pp. 617 -- 624.

\bibitem{yang2022sparse}
\BIBentryALTinterwordspacing
L.~Yang, R.~Ge, S.~Feng, and D.~Zhang, ``Learning projection views for sparse-view {CT} reconstruction,'' ser. MM '22.\hskip 1em plus 0.5em minus 0.4em\relax New York, NY, USA: Association for Computing Machinery, 2022, p. 2645–2653. [Online]. Available: \url{https://doi.org/10.1145/3503161.3548204}
\BIBentrySTDinterwordspacing

\bibitem{zhang2020metainv}
H.~Zhang, B.~Liu, H.~Yu, and B.~Dong, ``Metainv-net: Meta inversion network for sparse view ct image reconstruction,'' \emph{IEEE Transactions on Medical Imaging}, vol.~40, no.~2, pp. 621--634, 2020.

\bibitem{chen2017low}
H.~Chen, Y.~Zhang, M.~K. Kalra, F.~Lin, Y.~Chen, P.~Liao, J.~Zhou, and G.~Wang, ``Low-dose {CT} with a residual encoder-decoder convolutional neural network,'' \emph{IEEE Transactions on Medical Imaging}, vol.~36, no.~12, pp. 2524--2535, 2017.

\bibitem{yang2022inner}
L.~Yang, Z.~Li, R.~Ge, J.~Zhao, H.~Si, and D.~Zhang, ``Low-dose {CT} denoising via sinogram inner-structure transformer,'' \emph{IEEE Transactions on Medical Imaging}, pp. 1--1, 2022.

\bibitem{zhu2018image}
B.~Zhu, J.~Z. Liu, S.~F. Cauley, B.~R. Rosen, and M.~S. Rosen, ``Image reconstruction by domain-transform manifold learning,'' \emph{Nature}, vol. 555, no. 7697, pp. 487--492, 2018.

\bibitem{he2020radon}
J.~He, Y.~Wang, and J.~Ma, ``Radon inversion via deep learning,'' \emph{IEEE Transactions on Medical Imaging}, vol.~39, no.~6, pp. 2076--2087, 2020.

\bibitem{dhariwal2021diffusion}
P.~Dhariwal and A.~Nichol, ``Diffusion models beat gans on image synthesis,'' \emph{Advances in neural information processing systems}, vol.~34, pp. 8780--8794, 2021.

\bibitem{li2022diffusion}
X.~Li, J.~Thickstun, I.~Gulrajani, P.~S. Liang, and T.~B. Hashimoto, ``Diffusion-lm improves controllable text generation,'' \emph{Advances in Neural Information Processing Systems}, vol.~35, pp. 4328--4343, 2022.

\bibitem{lugmayr2022repaint}
A.~Lugmayr, M.~Danelljan, A.~Romero, F.~Yu, R.~Timofte, and L.~Van~Gool, ``Repaint: Inpainting using denoising diffusion probabilistic models,'' in \emph{Proceedings of the IEEE/CVF Conference on Computer Vision and Pattern Recognition}, 2022, pp. 11\,461--11\,471.

\bibitem{amit2021segdiff}
T.~Amit, T.~Shaharbany, E.~Nachmani, and L.~Wolf, ``Segdiff: Image segmentation with diffusion probabilistic models,'' \emph{arXiv preprint arXiv:2112.00390}, 2021.

\bibitem{zimmermann2021score}
R.~S. Zimmermann, L.~Schott, Y.~Song, B.~A. Dunn, and D.~A. Klindt, ``Score-based generative classifiers,'' in \emph{NeurIPS 2021 Workshop on Deep Generative Models and Downstream Applications}, 2021.

\bibitem{wolleb2022diffusion}
J.~Wolleb, F.~Bieder, R.~Sandk{\"u}hler, and P.~C. Cattin, ``Diffusion models for medical anomaly detection,'' in \emph{International Conference on Medical image computing and computer-assisted intervention}.\hskip 1em plus 0.5em minus 0.4em\relax Springer, 2022, pp. 35--45.

\bibitem{kim2022diffusemorph}
B.~Kim, I.~Han, and J.~C. Ye, ``Diffusemorph: Unsupervised deformable image registration using diffusion model,'' in \emph{European Conference on Computer Vision}.\hskip 1em plus 0.5em minus 0.4em\relax Springer, 2022, pp. 347--364.

\bibitem{packhauser2023generation}
K.~Packh{\"a}user, L.~Folle, F.~Thamm, and A.~Maier, ``Generation of anonymous chest radiographs using latent diffusion models for training thoracic abnormality classification systems,'' in \emph{2023 IEEE 20th International Symposium on Biomedical Imaging (ISBI)}.\hskip 1em plus 0.5em minus 0.4em\relax IEEE, 2023, pp. 1--5.

\bibitem{rombach2022high}
R.~Rombach, A.~Blattmann, D.~Lorenz, P.~Esser, and B.~Ommer, ``High-resolution image synthesis with latent diffusion models,'' in \emph{Proceedings of the IEEE/CVF Conference on Computer Vision and Pattern Recognition}, 2022, pp. 10\,684--10\,695.

\bibitem{meng2022novel}
X.~Meng, Y.~Gu, Y.~Pan, N.~Wang, P.~Xue, M.~Lu, X.~He, Y.~Zhan, and D.~Shen, ``A novel unified conditional score-based generative framework for multi-modal medical image completion,'' \emph{arXiv preprint arXiv:2207.03430}, 2022.

\bibitem{song2020score}
Y.~Song, J.~Sohl-Dickstein, D.~P. Kingma, A.~Kumar, S.~Ermon, and B.~Poole, ``Score-based generative modeling through stochastic differential equations,'' \emph{arXiv preprint arXiv:2011.13456}, 2020.

\bibitem{chung2022score}
H.~Chung and J.~C. Ye, ``Score-based diffusion models for accelerated mri,'' \emph{Medical image analysis}, vol.~80, p. 102479, 2022.

\bibitem{song2020denoising}
J.~Song, C.~Meng, and S.~Ermon, ``Denoising diffusion implicit models,'' \emph{arXiv preprint arXiv:2010.02502}, 2020.

\bibitem{liu2023dolce}
J.~Liu, R.~Anirudh, J.~J. Thiagarajan, S.~He, K.~A. Mohan, U.~S. Kamilov, and H.~Kim, ``Dolce: A model-based probabilistic diffusion framework for limited-angle {CT} reconstruction,'' in \emph{Proceedings of the IEEE/CVF International Conference on Computer Vision}, 2023, pp. 10\,498--10\,508.

\bibitem{bansal2024cold}
A.~Bansal, E.~Borgnia, H.-M. Chu, J.~Li, H.~Kazemi, F.~Huang, M.~Goldblum, J.~Geiping, and T.~Goldstein, ``Cold diffusion: Inverting arbitrary image transforms without noise,'' \emph{Advances in Neural Information Processing Systems}, vol.~36, 2024.

\bibitem{huang2023cdiffmr}
J.~Huang, A.~I. Aviles-Rivero, C.-B. Sch{\"o}nlieb, and G.~Yang, ``Cdiffmr: Can we replace the gaussian noise with k-space undersampling for fast mri?'' in \emph{International Conference on Medical Image Computing and Computer-Assisted Intervention MICCAI}.\hskip 1em plus 0.5em minus 0.4em\relax Springer, 2023, pp. 3--12.

\bibitem{moen2021low}
T.~R. Moen, B.~Chen, D.~R. Holmes~III, X.~Duan, Z.~Yu, L.~Yu, S.~Leng, J.~G. Fletcher, and C.~H. McCollough, ``Low-dose {CT} image and projection dataset,'' \emph{Medical Physics}, vol.~48, no.~2, pp. 902--911, 2021.

\bibitem{clark2013cancer}
K.~Clark, B.~Vendt, K.~Smith, J.~Freymann, J.~Kirby, P.~Koppel, S.~Moore, S.~Phillips, D.~Maffitt, M.~Pringle \emph{et~al.}, ``The cancer imaging archive (tcia): Maintaining and operating a public information repository,'' \emph{Journal of Digital Imaging}, vol.~26, no.~6, pp. 1045--1057, 2013.

\bibitem{Akin2016a}
O.~Akin, P.~Elnajjar, M.~Heller, and R.~Jarosz, ``The cancer genome atlas kidney renal clear cell carcinoma collection (tcga-kirc),'' in \emph{The Cancer Imaging Archive}, 2016.

\bibitem{he2016deep}
K.~He, X.~Zhang, S.~Ren, and J.~Sun, ``Deep residual learning for image recognition,'' in \emph{Proceedings of the IEEE Conference on Computer Vision and Pattern Recognition}, 2016, pp. 770--778.

\bibitem{torch_radon}
M.~Ronchetti, ``Torchradon: Fast differentiable routines for computed tomography,'' \emph{arXiv preprint arXiv:2009.14788}, 2020.

\bibitem{Zhang2018LPIPS}
R.~Zhang, P.~Isola, A.~A. Efros, E.~Shechtman, and O.~Wang, ``The unreasonable effectiveness of deep features as a perceptual metric,'' in \emph{Proceedings of the IEEE Conference on Computer Vision and Pattern Recognition}, 2018, pp. 586--595.

\bibitem{zhang2018sparse}
Z.~Zhang, X.~Liang, X.~Dong, Y.~Xie, and G.~Cao, ``A sparse-view {CT} reconstruction method based on combination of densenet and deconvolution,'' \emph{IEEE Transactions on Medical Imaging}, vol.~37, no.~6, pp. 1407--1417, 2018.

\bibitem{ma2023freeseed}
C.~Ma, Z.~Li, Y.~Zhang, J.~Zhang, and H.~Shan, ``Freeseed: Frequency-band-aware and self-guided network for sparse-view {CT} reconstruction,'' in \emph{Medical Image Computing and Computer Assisted Intervention MICCAI}, 2023.

\bibitem{liu2021swin}
Z.~Liu, Y.~Lin, Y.~Cao, H.~Hu, Y.~Wei, Z.~Zhang, S.~Lin, and B.~Guo, ``Swin transformer: Hierarchical vision transformer using shifted windows,'' in \emph{Proceedings of the IEEE/CVF International Conference on Computer Vision}, 2021, pp. 10\,012--10\,022.

\end{thebibliography}

\end{document}